\documentclass[10pt,twocolumn,letterpaper]{article}

\usepackage{iccv}
\usepackage{times}
\usepackage{epsfig}
\usepackage{graphicx}
\usepackage{amsmath}
\usepackage{amssymb}

\usepackage{fancyhdr}
\usepackage[pagebackref=true,breaklinks=true,letterpaper=true,colorlinks=false,bookmarks=false]{hyperref}

\iccvfinalcopy % *** Uncomment this line for the final submission

\def\iccvPaperID{7} % *** Enter the ICCV Paper ID here

\ificcvfinal\fi
\begin{document}

%%%%%%%%% TITLE
\title{Multi-Adapter RGBT Tracking}

\author{Chenglong Li$^{1,3}$, Andong Lu$^1$, Aihua Zheng$^1$, Zhengzheng Tu$^1$, Jin Tang$^{1,2}$\\
$^1$School of Computer Science and Technology, Anhui University, Hefei 230601, China\\
$^2$Key Laboratory of Industrial Image Processing and Analysis of Anhui Province\\
$^3$Institute of Physical Science and Information Technology, Anhui University, Hefei 230601, China\\
{\tt\small \{lcl1314, adlu\_ah\}@foxmail.com, ahzheng214@qq.com, zhengzhengahu@163.com, tangjin@ahu.edu.cn}
}

\maketitle
%\thispagestyle{empty}

%%%%%%%%% ABSTRACT
\begin{abstract}
The task of RGBT tracking aims to take the complementary advantages from visible spectrum and thermal infrared data to achieve robust visual tracking, and receives more and more attention in recent years.
Existing works focus on modality-specific information integration by introducing modality weights to achieve adaptive fusion or learning robust feature representations of different modalities.
%
%or modality-invariant feature learning by transforming different modalities into a same space.
%
Although these methods could effectively deploy the modality-specific properties, they ignore the potential values of modality-shared cues as well as instance-aware information, which are crucial for effective fusion of different modalities in RGBT tracking.
In this paper, we propose a novel Multi-Adapter convolutional Network (MANet) to jointly perform modality-shared, modality-specific and instance-aware feature learning in an end-to-end trained deep framework for RGBT tracking. 
We design three kinds of adapters 
%including the generality adapter, the modality adapter and the instance adapter 
within our network. 
In a specific, the generality adapter is to extract shared object representations, 
the modality adapter aims at encoding modality-specific information to deploy their complementary advantages, 
and the instance adapter is to model the appearance properties and temporal variations of a certain object. 
Moreover, to reduce computational complexity for real-time demand of visual tracking, we design a parallel structure of generic adapter and modality adapter. 
%
%We also develop a progressive strategy to train the whole network effectively.
%
Extensive experiments on two RGBT tracking benchmark datasets demonstrate the outstanding performance of the proposed tracker against other state-of-the-art RGB and RGBT tracking algorithms.
\end{abstract}

%%%%%%%%% BODY TEXT

%-------------------------------------------------------------------------
\section{Introduction}
The problem of RGBT tracking could be considered as an extension of visual tracking, 
and its goal is to estimate target states using the complementary advantages of visible spectrum (called RGB in this paper) and thermal infrared information given the initial state in the first pair of frame. 
It has been receiving much more attention recently and becoming more and more popular partly due to the following reasons: 
i) RGB and thermal data have strong complementary advantages and thus could overcome imaging limitations of individual source~\cite{Li17weld,Wu17iccv,Xu17cvpr,Li17rgbt210}. 
ii) Thermal infrared cameras are economically available in recent years~\cite{Gade14mva}, making RGBT data easier to access in various applications, such as object segmentation~\cite{Li17weld}, person Re-ID~\cite{Wu17iccv} and pedestrian detection~\cite{Hwang15cvpr,Xu17cvpr}. 
iii) Recent RGBT tracking benchmark datasets~\cite{Li16tip,Li17rgbt210} provide a flexible evaluation platform of various RGBT trackers. 
iv) The VOT2019 challenge has announced ``VOT-RGBT challenge'' to address short-term trackers that use RGB and thermal infrared modalities\footnote{http://www.votchallenge.net/}.
Although much progress has been achieved, how to make the best of RGB and thermal information for robust RGBT tracking is an open problem.

Existing works focus on modality-specific information integration from two major aspects. 
One is to introduce modality weights that reflect their reliabilities in tracking prediction to achieve adaptive fusion of different modalities. 
For example, Li et al.~\cite{Li16tip} integrate computation of modality weights and sparse representation in a joint model and perform online object tracking in Bayesian filtering framework.
Lan et al.~\cite{lan18aaai} learn modality weights and classifiers of different modalities in a max-margin learning framework.
Another is to learn robust feature representations of different modalities.
For example, Li~\emph{et al.}~\cite{Li17rgbt210} propose to represent target object using a collaborative graph with local image patches as nodes. 
They learn a patch-based weighted RGBT features to fuse different modalities and alleviate background effects of bounding box descriptions simultaneously.
Li~\emph{et al.}~\cite{Li18eccv} propose a cross-modal ranking algorithm to improve the robustness of weight computation which yields a more robust RGBT feature representations.
In addition, there are some methods to take both of above two aspects into considerations.
For example, Zhu~\emph{et al.}.~\cite{Zhu18FANet} propose a feature aggregation network, which includes a hierarchical aggregation module to learn robust features from each modality and a quality-aware fusion module to achieve weight-based integration.

These methods could effectively deploy the modality-specific properties, but most of them ignore the potential values of modality-shared cues as well as instance-aware information, which are crucial for effective fusion of different modalities in RGBT tracking.

In this paper, we propose a novel Multi-Adapter convolutional Network (MANet) to jointly perform modality-shared, modality-specific and instance-aware feature learning in an end-to-end trained deep framework for RGBT tracking. 
We design three kinds of adapters including the generality adapter, the modality adapter and the instance adapter 
within our network. 

Since existing RGBT tracking methods ignore modality-shared cues as discussed above, we design a generality adapter to extract shared object representations.
Visible spectrum and thermal infrared data are captured from cameras of different imaging bands, and thus reflect different properties of target objects.
In spite of it, they share some common information like object boundaries and some fine-grained textures, and thus how to model them plays a critical role in learning collaborative representations of different modalities.
Based on the observation, we employ the generality adapter to extract shared object representations across different modalities.
To this end, we adopt the first three convolutional layers of VGGNet-M~\cite{vgg15iclr} as the generality adapter. 
It should be noted that other networks like Inception Network~\cite{InceptionNet15icml} and Residue neural Network (ResNet)~\cite{ResNet} could also be applied in our framework.
We adopt VGGNet-M for its good balance of accuracy and complexity.

To model characteristics of each modality and make best use of the complementary advantages from RGB and thermal modalities, we need to design a subnetwork to learn modality-specific feature representations. 
Existing works~\cite{li18nuecom,Zhu18FANet} often develop a two-stream Convolutional Neural Network (CNN) to extract RGB and thermal features respectively.
The two-stream CNNs usually include many parameters and thus might degrade tracking efficiency.
To reduce computational complexity for real-time demand of visual tracking, we propose a modality adapter, whose structure is similar to ResNet~\cite{ResNet},
to effectively extract modality-specific feature representations with much less computational burden.
In particular, we design a parallel network structure that includes a small convolution kernel (e.g., 3$\times$3 or 1$\times$1) at each convolutional layer of generic adapter.
% explain
Although only small convolutional kernels are used, our modality adapter is able to encode modality-specific information effectively.
The reason is that different modalities could share a larger portion of their parameters as discussed above therefore the number of the modality-specific parameters should be much smaller than the generality adapter. 

Individual instance objects involve different class labels, moving
patterns, and appearances, 
thus tracking algorithms suffer from instance-specific challenges such as occlusion, deformation, motion blur.
Furthermore, even for a certain instance object, its appearance might vary a lot so that tracking models trained in the initial frames are ineffective to track.
Motivated by Multi-Domain Convolutional Network (MDNet)~\cite{MDNet15cvpr}, we further design a instance adapter to model the appearance properties and temporal variations of a certain object.
The instance adapter is composed of three fully connected layers, and we adopt offline and online learning strategies to capture appearance properties and temporal variations of a certain object respectively.
%
%The proposed modality adapter would perform effective fusion of different modalities and thus significantly improve the tracking performance. In addition, motivated by, we design a parallel network structure to reduce computational burden of modality adapter while preserving its high accuracy. The advances of our network architecture over typical ones are shown in Fig.~\ref{fig::structures}.

Considering the network learning of each modality as an individual task, the joint learning of our MANet is essentially formulated as a multi-task learning problem. 
Therefore, we develop an effective progressive learning algorithm to train our MANet.
In particular, the generality adapter is first pre-trained on large-scale image classification dataset, and then fine-tuned on RGBT dataset.
By fixing the parameters of generality adapter, we train the modality adapters using RGBT dataset.
For the training of the instance adapter, we adopt offline and online algorithms
to learn its parameters as shown in MDNet~\cite{MDNet15cvpr}.
Note that the instance adapter is always trained in whole process except for pre-train stage, and thus when training instance adapters, the whole network is end-to-end trained.
Finally, we perform the online tracking by evaluating the candidate regions randomly sampled around the previous tracking result. 
Extensive experiments on two RGBT tracking benchmark datasets demonstrate the outstanding performance of the proposed tracker.

This paper makes the following major contributions to RGBT tracking and related applications.

\begin{itemize}
\item It presents a novel end-to-end trained deep network MANet for RGBT tracking. 
MANet consists of three types of adapters to provide a powerful RGBT deep representations to well handle various challenges in RGBT tracking.
Our MANet is generic and could handle multiple modalities with the number larger than two.
We will release the code of MANet to public for reproducible research. 

\item It presents an effective parallel structure of generality and modality adapters to reduce computational complexity for real-time demand of RGBT tracking. 
%Comparing with serial structures, our parallel adapters are able to reduce computational complexity.
%
The method of our parallel design is scalable, and could be extended to more branches for other applications, such as category-aware and challenge-aware adapters.

\item Extensive experiments on two RGBT tracking benchmark datasets suggest that the proposed tracker achieves the outstanding performance and yields a new state-of-the-art for RGBT tracking.

\end{itemize}

\section{Related Work}

\subsection{RGBT Tracking Methods}
RGBT tracking receives much attention recently and becomes more and more popular~\cite{Li16tip,Li17tsmcs,Li17rgbt210,lan18aaai,li18nuecom,Li18eccv}. 
Recent works~\cite{Li17tsmcs,Li16tip,lan18aaai} employ reconstruction residues~\cite{Li17tsmcs,Li16tip} or classification scores~\cite{lan18aaai} to guide the weights learning of modalities to achieve adaptive fusion of RGB and thermal modalities. 
However, these methods tend to lose target objects in tracking process when the reconstruction residues or classification scores are unreliable in representing modal reliabilities.

Recent studies are focusing on the construction of robust RGBT feature representations~\cite{Li17rgbt210,li18nuecom,Li18eccv}. 
Li \emph{et al.}~\cite{Li17rgbt210} propose a graph learning approach based on the weighted sparse representation to construct a patch-based RGBT feature descriptor, and perform online tracking via the structured SVM. 
Li \emph{et al.}~\cite{Li18eccv} propose a cross-modal ranking algorithm, which takes both heterogeneous properties between RGB and thermal infrared modalities and noise effects of initial ranking seeds into account. The ranking results are used as patch weights to construct robust RGBT feature representations. 
To adaptively fuse RGB and thermal infrared modalities, Li \emph{et al.}~\cite{li18nuecom} propose to select most discriminative feature maps in a two-stream convolutional neural network. 
These methods rely on either handcrafted features or single-adapter deep structures to localize objects, and might be difficult to handle the challenges of significant appearance changes caused by deformation, abrupt motion, background clutter and occlusion, \emph{etc}.

\subsection{Multi-Domain Tracking Methods}
Nam \emph{et al.}~\cite{MDNet15cvpr} propose the Multi-Domain Network (MDNet), which achieves outstanding performance in visual tracking and win the VOT2015 challenge. 
MDNet uses a CNN-based backbone pretrained offline to extract generic target representations, and the fully connected layers updated online to adapt temporal variations of target objects. 
In MDNet, each domain corresponds to one video sequence. 
Due to its effectiveness in visual tracking, extensive works~\cite{BranchOut17cvpr,MetaTracker18eccv,RT-MDNet18eccv} are developed based on MDNet. 
For example, Park and Berg~\cite{MetaTracker18eccv} introduce the meta-learning to MDNet to adjust the initial parameters of deep networks, which could quickly adapt to temporal variations of target objects in future frames. 
To improve the efficiency of MDNet, Jung~\emph{et al.}~\cite{RT-MDNet18eccv} propose an improved RoIAlign operation to extract more accurate representations directly on feature maps for targets.
Except for visual tracking task, to tackle a large variety of different problems within a single model, Rebuffi \emph{et al.}~\cite{Rebuffi2017nips,Rebuffi2018cvpr} introduce a design for multivalent neural network architectures for multiple-domain learning.
Different from these structures, we propose the multi-adapter deep structures to learn more powerful RGBT feature representations, 
in which three types of adapters are designed to learn task-specific representations. Furthermore, we propose a parallel deep architecture to reduce computational burden effectively.

\begin{figure*}[t]
\centering
\includegraphics[width=2.1\columnwidth]{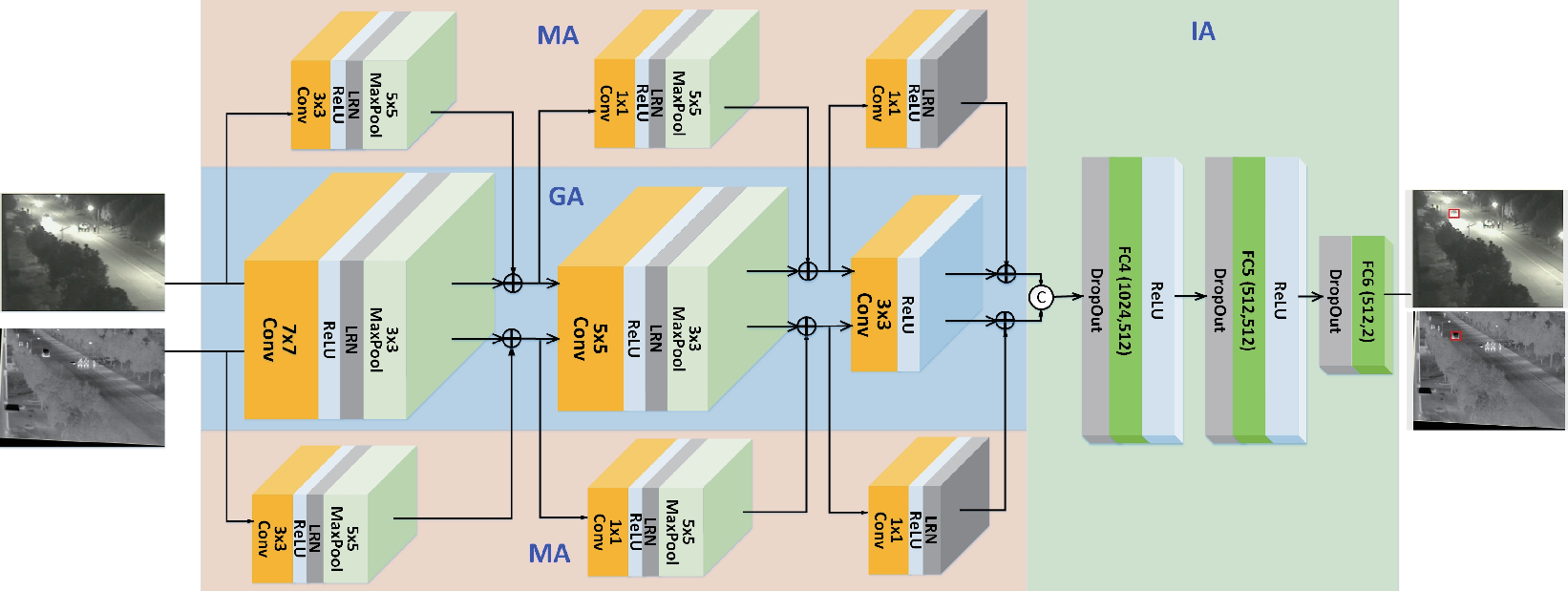}
\caption{Pipeline of MANet. Herein, + and C denote the operations of addition and concatenation respectively. ReLU and LRN refer to the rectified linear unit and the local response normalization unit respectively. The blue, pink and green blocks represent the generic adapter (GA), modality adapter (MA) and instance adapter (IA), respectively.}
\label{fig::pipeline}
\end{figure*}

\section{Multi-Adapter Convolutional Network}
In this section, we will elaborate the proposed multi-adapter network (MANet), including network architecture and corresponding learning algorithm.

\subsection{Network Architecture}
The pipeline of MANet is shown in Fig.~\ref{fig::pipeline}, in which the detailed parameter settings are presented. 
Our MANet consists of three kinds network blocks, i.e., generality adapter (GA), modality adapter (MA), instance adapter (IA).

{\flushleft {\bf Generality adapter (GA)}}.
Visible spectrum and thermal infrared data are captured from cameras of different imaging bands, and thus reflect different properties of target objects.
In spite of it, they share some common information like object boundaries and some fine-grained textures, and thus how to model them plays a critical role in learning collaborative representations of different modalities.
However, existing works~\cite{li18nuecom,Zhu18FANet} usually model different modalities separately, and thus ignore modality-shared features. 
Furthermore, separate processing for each modality would introduce a lot of redundant parameters as different modalities should have a large portion of shared parameters.
To handle these problems, we design a generality adapter (GA) to extract shared object representations across different modalities.
There are many potential networks~\cite{vgg15iclr,ResNet} to be used for our GA, and we select the VGG-M network~\cite{vgg15iclr} for its good balance trade-off of the effectiveness and efficiency.
In a specific, our GA consists of the first three layers intercepted from the VGG-M network, where the convolution kernel sizes are $7\times 7\times 96$, $5\times 5\times 256$, $3\times 3\times 512$ respectively. 
Each layer of GA is composed of a convolution layer, a activation function of rectified linear unit (ReLU), local response normalization (LRN), and a max pooling layer. The details are shown in Fig.~\ref{fig::pipeline}.

{\flushleft {\bf Modality adapter (MA)}}.
The RGB and thermal modalities are heterogeneous with different properties, and thus only using GA is insufficient for RGBT feature presentations.
To model characteristics of each modality and make best use of the complementary advantages from RGB and thermal modalities, we need to design a subnetwork to learn modality-specific feature representations. 
Existing works~\cite{li18nuecom,Zhu18FANet} often develop a two-stream Convolutional Neural Network (CNN) to extract RGB and thermal features respectively.
The two-stream CNNs ignore modality-shared feature learning discussed abover and also usually include many parameters, which might degrade tracking accuracy and efficiency respectively. 
To improve RGBT feature representations and reduce computational complexity for real-time demand of visual tracking, we propose a modality adapter (MA) that is built on GA to effectively extract modality-specific feature representations with a little computational burden.

In a specific, we design a parallel network structure that includes a small convolution kernel (e.g., 3$\times$3 or 1$\times$1) at each convolutional layer of GA.
% explain
Although only small convolutional kernels are used, our MA is able to encode modality-specific information effectively, 
since different modalities should share a larger portion of their parameters as discussed above so that the number of the modality-specific parameters should much smaller than GA.
The experimental results also demonstrate the effectiveness of our settings.
In particular, we develop an adaptive scheme to determine the size of convolution kernel of GA according to the kernel size of GA. 
Therefore, the kernel sizes of our MA are set to 3$\times$3 (7$\times$7 in GA), 1$\times$1 (5$\times$5) and 1$\times$1 (3$\times$3) respectively. 
Furthermore, followed by a convolution, each layer of MA also includes the ReLU activation function, LRN and max pooling for more effective representations.  
The details are shown in Fig.~\ref{fig::pipeline}.

Taking the thermal MA as an example, we introduce the details of the combination of MA and GA.
The network parameters of GA and thermal MA are denoted as ${\bf W}^{GA}$ and ${\bf W}^{MA}$ respectively, and the input of thermal modality is denoted as ${\bf T}$. 
The output of a layer in MANet is ${\bf F_T}$ which is formulated as follows:
\begin{equation}
\begin{aligned}
&{\bf F_T}={\bf W}^{GA}\ast{\bf T}+{\bf W}^{MA}\ast{\bf T}\\
\end{aligned}
\label{eq::1}
\end{equation}
where $\ast$ represents a convolution operation, and ${\bf W}^{GA}$ is with the size of $L\times L$.

Next, we will explain why our MA defined in~\eqref{eq::1} can capture modality-specific feature representations effectively and efficiently.
We first design an operation ${\bf diag}_L({\bf W}^{MA})$ to the matrix ${\bf W}^{MA}$ into a new matrix with the size of $L\times L$ that is the same with GA:
\begin{equation}
\begin{aligned}
&{\bf diag}_L({{\bf W}}^{MA})_{wh}=
\left\{
            \begin{array}{lcl}
             {\bf W}^{MA}_{ij},w=\frac{\emph{L}-a}{2}+i,h=\frac{\emph{L}-b}{2}+j. \\
			\qquad\qquad{s.t. }{0<i<a,0<j<b.} \\             
             \\
             \qquad{\bf 0},\qquad {\emph{otherwise.} }
            \end{array}              
        \right.    
\end{aligned}
\label{eq::2}
\end{equation}
where $a\times b$ denotes the size of ${\bf W}^{MA}$.
In this way, we can merge GA and MA in~\eqref{eq::1} as follows:
\begin{equation}
\centering
\begin{aligned}
&{\bf F_T}=({\bf W}^{GA}+{\bf diag}_L({\bf W}^{MA}))\ast{\bf T}\triangleq {\bf M}\ast{\bf T},
\end{aligned}
\label{eq::3}
\end{equation}
where ${\bf M}={\bf W}^{GA}+{\bf diag}_L({\bf W}^{MA})$.

From~\eqref{eq::3} we can see that MA and GA can be fused by computing new weight matrix M explicitly, which does not introduce extra computing costs during training and tracking phases.
Meanwhile, we can control the adaptability of our MANet to the modality by adjusting the network parameters of modality adapter, i.e., ${\bf W}^{MA}$.
Therefore, our MA with the parallel structure can learn modality-specific features in the training process.

{\flushleft {\bf Instance adapter (IA)}}.
Individual instance objects involve different class labels, moving
patterns, and different appearances, 
and tracking algorithms suffer from instance-specific challenges such as occlusion, deformation, motion blur.
Furthermore, even for a instance object, its appearance might vary a lot so that tracking models trained in the initial frames are ineffective to track.
To handle these problems, we integrate a instance adapter (IA) inspired by MDNet~\cite{MDNet15cvpr} to model the appearance properties and temporal variations of a certain object.
In a specific, IA is composed of three fully connected (FC) layers with dropout layers named as FC4, FC5 and FC6 whose output dimensions are 512, 512, 2 respectively. 
The ReLU activation function is followed by the FC4 and FC5 layers, and the FC6 layer is with the softmax cross-entropy loss as a binary classification layer. 

\subsection{Progressive Learning Algorithm}
Considering the network learning of each modality as an
individual task, the joint learning of our MANet is essentially formulated as a multi-task learning problem. 
Therefore, to train our MANet effectively,
we develop an effective progressive learning algorithm which is an effective solver on the problems like multi-task learning.

{\flushleft {\bf GA Training}}.
We initialize parameters of our GA using the pre-trained model in VGG-M~\cite{vgg15iclr}, 
and then fine-tune it using RGBT dataset. 
Note that when we conduct testing on the GTOT dataset~\cite{Li16tip}, we fine-tune GA using the RGBT234 dataset~\cite{Li18arXiv}, and vice versa. 
We use the stochastic gradient descent (SGD) algorithm~\cite{Ketkar2014Stochastic} to train GA, and set the learning parameters as follows.
The learning rates of all convolutional layers are set to 0.0001, and the learning rates of all fully connected layers are set to 0.001. 
The number of epochs is set to 100. 
In this stage, we only save the parameters of GA and discard parameters of MA and IA.

{\flushleft {\bf MA Training}}.
In the second stage, we aim to learn the parameters of MA. 
We first load the parameters of GA while fixing them in training. 
We employ the SGD algorithm to train MA and set the learning parameters as follows.
The learning rates of convolutional layers and FC layers are set to 0.0001 and 0.0005, respectively. 
The number of epochs is set to 100. 
In this stage, we save the parameters of MA and FC4-5 layers, and discard parameters of FC6.

{\flushleft {\bf IA Training}}.
For IA, we adopt both offline and online fashions to learn its parameters. 
The former is used to capture the characteristics of target instances, and discards parameters of last FC layer during the training. 
The latter is used to capture temporal dynamics of target appearance, and a new last FC layer is learned for a new instance in first frame and last three FC layers are updated in subsequent frames with several frames interval.
It should be noted that IA is always trained in whole process except for pre-train stage,
and thus when training IA, the whole network is end-to-end trained.

\subsection{Implementation}
In the implementation, we collected 500 positive samples (IoU with ground truth greater than 0.7) and 5000 negative samples (IoU with ground truth less than 0.5) as the training samples in the first frame to learn parameters of IA (FC4-5-6), where the learning rate of FC4-5 is set to 0.0001, and the learning rate of FC6 is set to 0.001. 
The training iteration is set to 30. 
In subsequent frames, we draw positive (IoU with ground truth greater than 0.7) and negative samples (IoU with ground truth less than 0.3) as training samples for short-term update and long-term update to train IA~\cite{MDNet15cvpr}. 
The learning rates of FC4-5 and FC6 are set to 0.0002 and 0.002, respectively.

\section{Online Tracking}
During tracking process, we fixed all parameters of generality and modality adapters. 
We replace the instance adapter with a new one to fit the target instance in a new RGBT video sequence. 
At time ${\bf t}$, we take Gaussian sampling centered on previous tracking result ${\bf X}_{t-1}$ at time ${\bf t-1}$, and collect 256 candidate regions as ${{\bf x}^i_t}$. 
We use these candidate regions as inputs to our network and obtain their classification scores. 
The positive and negative scores for each sample are denoted as $f^+(x^i_t)$ and $f^-(x^i_t)$, respectively. 
We select the candidate region sample with the highest score as the tracking result ${\bf X}_t^\ast$ at time ${\bf t}$, and the formula expression is as follows:
\begin{equation}
{\bf X}_t^\ast=\mathop{ \arg\max}_{i=0,...,255}\ \mathrm{\emph{f}^+({\bf x}^i_t)}
\label{eq::6} 
\end{equation}

Followed by MDNet~\cite{MDNet15cvpr}, we use the bounding box regression technique to improve the problem of target scale transformation in the tracking process and improve the accuracy of positioning. 
It is worth noting that we only train it in the first frame for tracking efficiency.

\begin{figure}[t]
\centering
\includegraphics[width=1\columnwidth]{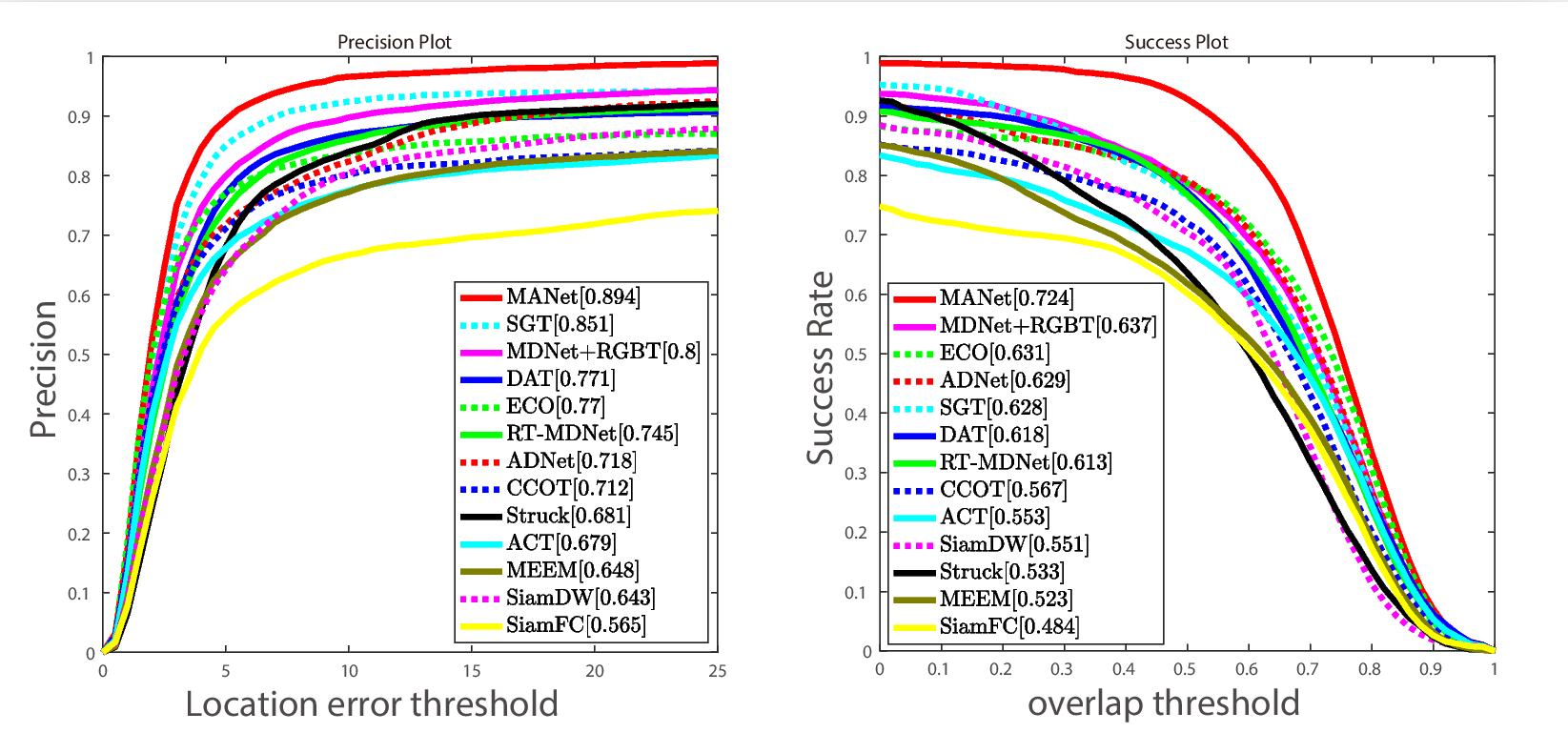}
\centering
\caption{PR and SR curves of different tracking result on GTOT dataset, where the representative PR and SR scores are presented in the legend.}
\label{fig:GTOT_results}
\end{figure}

\begin{figure}[t]
\centering
\includegraphics[width=\columnwidth]{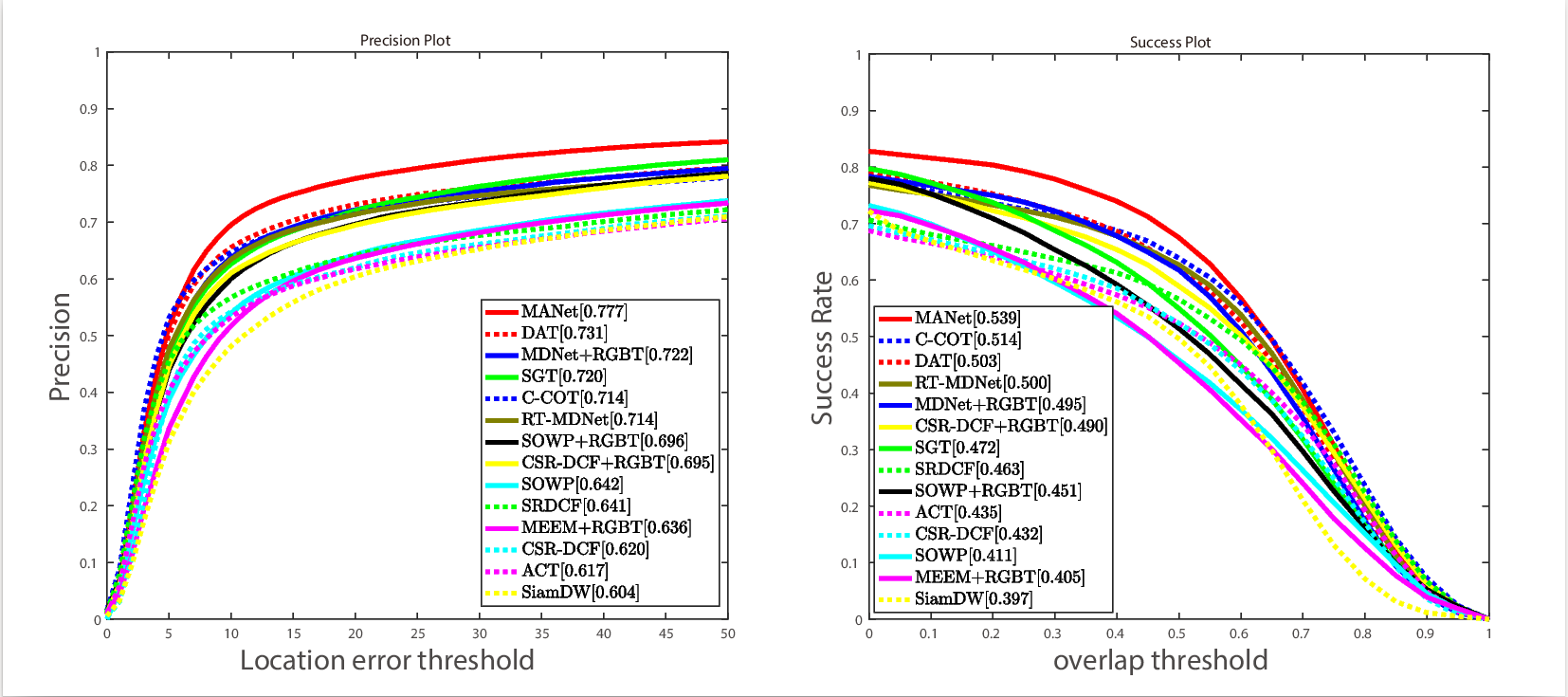}
\centering
\caption{PR and SR curves of different tracking result on RGBT234 dataset, where the representative PR and SR scores are presented in the legend.}
\label{fig:RGBT234_results}
\end{figure}

\section{Performance Evaluation}
In this section, we will compare our MANet with state-of-the-art RGB and RGBT tracking methods on two RGBT tracking benchmark datasets, GTOT~\cite{Li16tip} and RGBT234~\cite{Li18arXiv}, 
and then evaluate the main components of MANet in detail for better understanding of our approach.

\subsection{Evaluation Data and Metrics}
The GTOT dataset~\cite{Li16tip} contains 50 spatially and temporally aligned pairs of RGB and thermal infrared video sequences. 
It includes seven challenging factors for comprehensive evaluation of different tracking algorithms. 
The RGBT234 dataset is a large-scale RGBT tracking dataset extended from the RGBT210 dataset~\cite{Li17rgbt210}. 
It contains a total of 234 high-aligned pairs of RGB and thermal infrared video sequences, and has about 200,000 frames and the longest video sequence reaches 4,000 frames. 
Note that appearance of target objects in this dataset is significantly changing over time caused by occlusion, motion blur, camera moving and illumination challenges, and it is thus challenging enough to evaluate different trackers.

We employ the widely used tracking evaluation metrics including precision rate (PR) and success rate (SR) for quantitative performance evaluation. 
In a specific, PR is the percentage of frames whose output location is within a threshold distance of the ground truth, and we compute the representative PR score by setting the threshold to be 5 and 20 pixels for GTOT and RGBT234 datasets respectively (since the target object in GTOT is generally small). 
SR is the percentage of the frames whose overlap ratio between the output bounding box and the ground truth bounding box is larger than a threshold, and we compute the representative SR score by the area under the curves.

\subsection{Evaluation on GTOT dataset}
On the GTOT dataset, we compare with 12 trackers, including SGT~\cite{Li17rgbt210}, MDNet~\cite{MDNet15cvpr}+RGBT, Struck~\cite{Stuck11iccv}+RGBT, DAT~\cite{Pu2018Deep}, ECO~\cite{ECO17cvpr}, RT-MDNet~\cite{Jung2018Real}, ADNet~\cite{ADNet17cvpr}, C-COT~\cite{C-COT16eccv}, SiamDW~\cite{Zhipeng2019Deeper}, ACT~\cite{Chen2018Real}, MEEM~\cite{MEEM14eccv} and SiamFC~\cite{Bertinetto2016Fully}, where the first three methods are RGBT-based trackers and the remaining are RGB-based. 
Herein, we extend some RGB tracking methods to RGB-T ones for fair comparison by concatenating RGB and thermal features into a single vector or regarding the thermal as an extra channel.

%\begin{figure*}[t]
%\centering
%\includegraphics[width=2.1\columnwidth]{images/Challenges_PR}
%\centering
%\caption{ PR evaluation results on various challenges comparing to the-state-of-the-art methods on RGBT234
%.}
%\label{fig::Challenges_PR}
%\end{figure*}

\begin{figure*}[t]
\centering
\includegraphics[width=2.1\columnwidth]{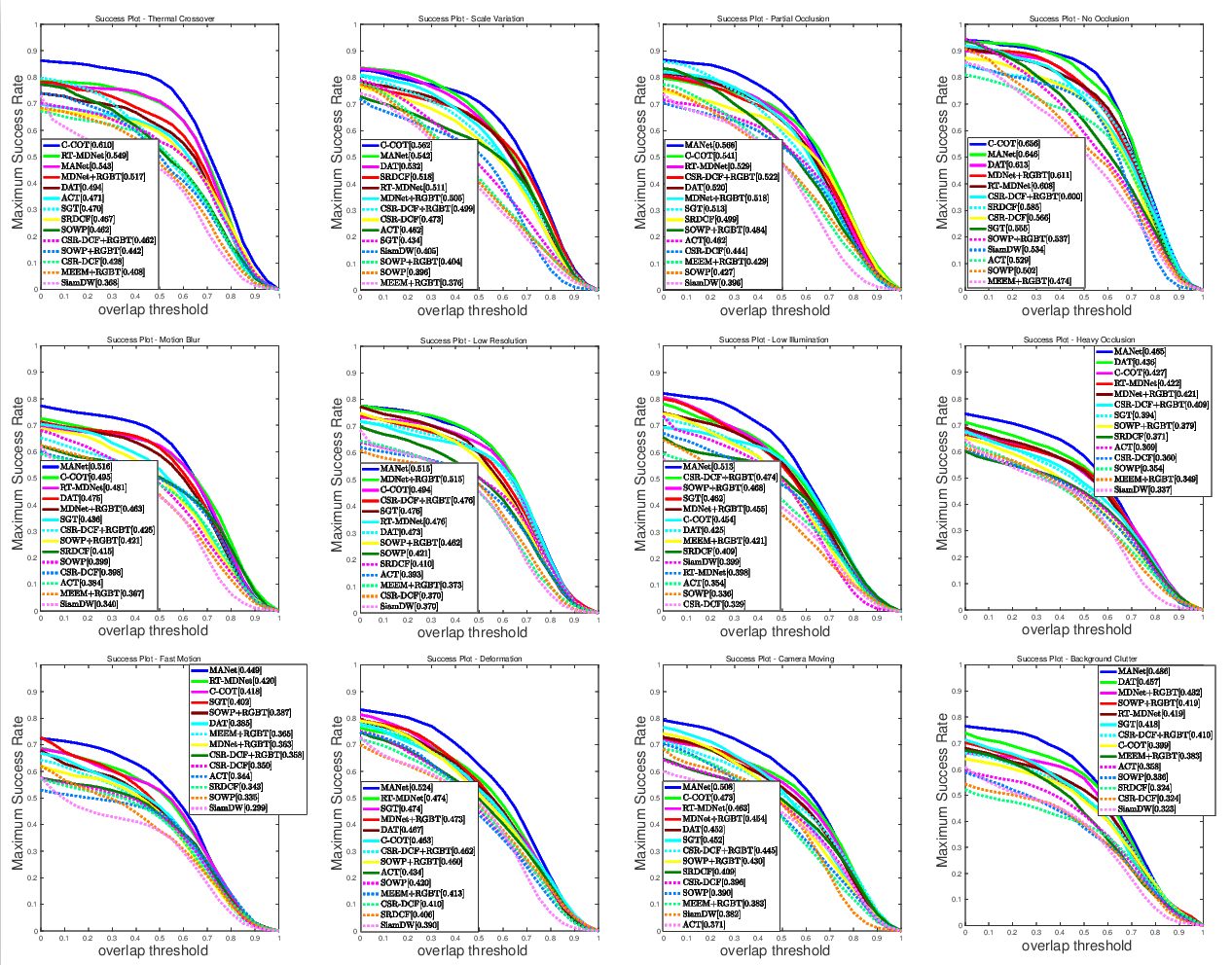}
\centering
\caption{SR evaluation results on various challenges comparing to the-state-of-the-art methods on RGBT234.}
\label{fig::Challenges_SR}
\end{figure*}

The evaluation results are reported in Fig.~\ref{fig:GTOT_results}. 
From the results, we can see that our MANet significantly outperforms the other trackers on GTOT. 
In particular, MANet (89.4\%/72.4\% in PR/SR) outperforms 4.3\% over the second best tracker SGT ~\cite{Li17rgbt210}(85.1\%) in PR, and 8.7\% over MDNet~\cite{MDNet15cvpr}+RGBT (63.7\%) in SR. 
It demonstrates the effectiveness of the multiple adapters for RGBT target representations. 
In addition, the remarkable superior performance over the state-of-the-art trackers like SiamDW~\cite{Zhipeng2019Deeper}, C-COT ~\cite{C-COT16eccv}and ECO~\cite{ECO17cvpr} suggests that our method is able to make best use of thermal modalities to boost tracking performance.

\subsection{Evaluation on RGBT234 dataset}
To further demonstrate the effectiveness of our method, we conduct the experiments on a larger dataset, RGBT234 dataset~\cite{li18nuecom}, and evaluate both the overall and challenge-based performance for comprehensive comparison.

{\flushleft \bf{Overall performance}}.
To evalute the overall performance, we compare our method with 13 baseline trackers, including SGT~\cite{Li17rgbt210}, 
MDNet~\cite{MDNet15cvpr}+RGBT, SOWP~\cite{Kim15iccv}+RGBT, CSR-DCF~\cite{dcf-csr16cvpr}+RGBT,MEEM
 \cite{MEEM14eccv}+RGBT, DAT~\cite{Pu2018Deep},  RT-MDNet~\cite{Jung2018Real}, SOWP~\cite{Kim15iccv}, 
C-COT~\cite{C-COT16eccv}, SiamDW~\cite{Zhipeng2019Deeper}, ACT~\cite{Chen2018Real}, CSR-DCF and 
SRDCF~\cite {danelljan2015learning}, where the first five methods are RGBT-based trackers and the remaining are RGB-based.

\begin{figure*}[t]
\centering
\includegraphics[width=2.1\columnwidth]{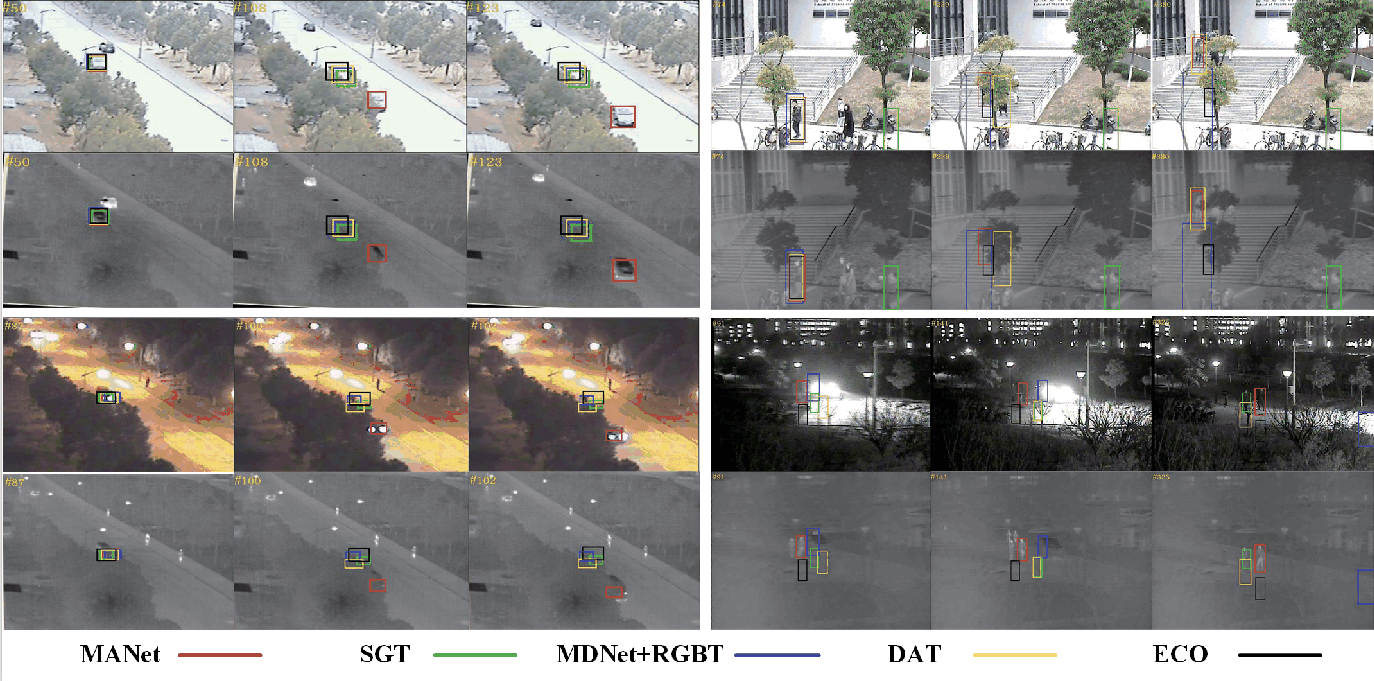}
\centering
\caption{Visual examples of our tracker comparing with four state-of-the-art baseline methods.}
\label{fig:Visual_examples}
\end{figure*}

From the results in Fig.~\ref{fig:RGBT234_results}, we can see that our MANet achieves superior performance over all the other trackers on RGBT234, further justifying the effectiveness of our MANet. 
In particular, MANet (77.7\%/53.9\% in PR/SR) outperforms 4.6\% over the second best tracker DAT (73.1\%)  in PR, and 2.5\% over C-COT (51.4\%)in SR. 
We also present the qualitative results against SGT, MDNet+RGBT, DAT and ECO in Fig.~\ref{fig:Visual_examples}, which visually demonstrates the effectiveness of our approach.

{\flushleft \bf{Challenge-based performance}}.
The annotated challenges in RGBT234 include, no occlusion (NO), partial occlusion (PO), heavy occlusion (HO), low illumination (LI), low resolution (LR), thermal crossover (TC), deformation (DEF), fast motion (FM), scale variation (SV), motion blur (MB), camera moving (CM) and background clutter (BC). 
The evaluation results are shown in Fig.~\ref{fig::Challenges_SR}.

From the results, MANet beats the other methods in most of challenges, especially in the challenges of partial occlusion (PO), heavy occlusion (HO), low illumination(LI), fast motion(FM), deformation (DEF), camera moving (CM) and background clutter (BC). 
It justifies the effectiveness of the proposed approach in handling above challenges. 
However, MANet performs overshadowed in the challenges of thermal crossover (TC) and no occlusion (NO). 
In a specific, thermal crossover makes thermal information unreliable to track targets, and fusing thermal modality might affect tracking performance.
Therefore, RGB tracker C-COT ~\cite{C-COT16eccv} performs well against our MANet. 
We can alleviate this issue by introducing modality weights to mitigate effects of noisy modality, and leave it in future work. 
For the challenge of no occlusion, these video sequences are less challenging, and thus RGB information might be enough to track targets. 
In addition, our MANet is comparable with the best RGB tracker C-COT~\cite{C-COT16eccv}.

\subsection{Ablation Study}
To validate the effectiveness of our major components, we implement three variants, including 
1) MANet-NO-MA, that removes modality adapter in tracking, 
2) MANet-NO-IA, that removes online update of instance adapter in tracking, and 
3) MANet-NO-GA, that removes generality adapter in tracking.

\begin{figure}[t]
\centering
\includegraphics[width=\columnwidth]{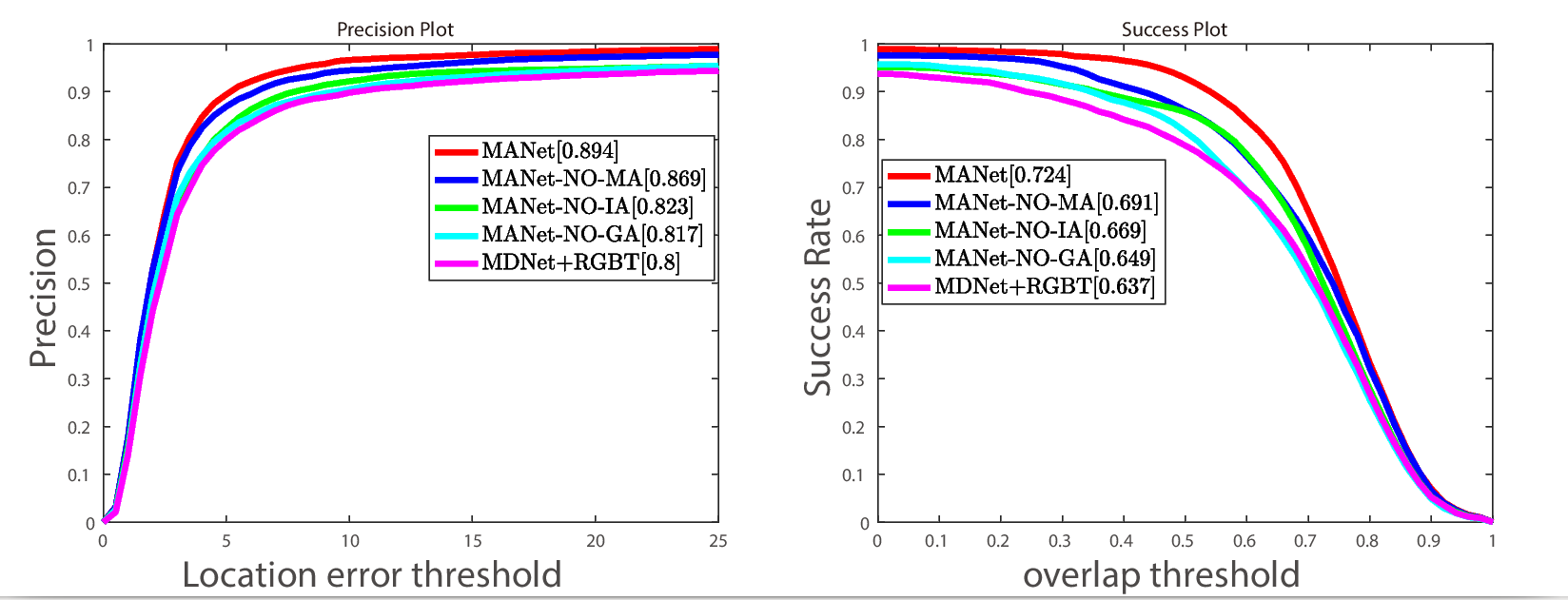}
\centering
\caption{Comparison results of MANet and its variants on GTOT dataset, where the representative PR and SR scores are presented in the legend.}
\label{fig:GTOT_component}
\end{figure}

The comparison results on GTOT are shown in Fig.~\ref{fig:GTOT_component}. 
From the results, we can make the following observations and conclusions. 
1) MANet is superior over MANet-NO-MA, which suggests the proposed modality adapter is helpful to capture modality-specific properties and thus improve tracking performance. 
2) MANet outperforms MANet-NO-IA with a clear margin. 
It specifies the importance of the adaptation to temporal variations of target objects. 
3) The large superior performance of MANet over MANet-NO-GA demonstrates the effectiveness of generic representations of target objects.

\subsection{Efficiency Analysis}
We implement our approach on the PyTorch platform with 3.75 GHz Intel Core I7-7700K, NVIDIA GeForce GTX 1080 GPU and 16G RAM. 
The frames rates of MANet, MANet-NO-MA, and MDNet~\cite{MDNet15cvpr}+RGBT are 1.11 FPS, 1.43 FPS and 1.61 FPS, respectively. 
Herein, MDNet+RGBT is implemented by regarding thermal image as an extra channel of RGB image and then running MDNet~\cite{MDNet15cvpr} to obtain tracking results. 
Note that the generality adapter of our MANet shares common weights of RGB and thermal sources, and the computations of generic and modality-specific adapters are parallel. 
Therefore, MANet does not bring much computational burden against MDNet-NO-MA and also against MDNet+RGBT, but significantly outperforms them in both PR and SR.

\section{Conclusion}
In this paper, we have proposed a powerful RGBT representation method based on the multi-adapter network MANet for visual tracking. 
MANet includes three types of adapters to extract generic, modality-specific and instance-aware deep feature representations respectively, and thus can well represent the target objects to address various tracking challenges. 
In addition, we design a parallel structure to reduce computational burden effectively. 
Extensive experiments on two benchmark datasets demonstrate the effectiveness and efficiency of the proposed tracking method. 
In future work, we will extend our framework to handle more modalities like depth to further boost the performance of tracking, and improve network structure to achieve real-time performance like RT-MDNet~\cite{RT-MDNet18eccv}.

{\small
\bibliographystyle{ieee}
\bibliography{mybibfiles}
}

\end{document}